\documentclass[10pt,twocolumn,letterpaper]{article}

\usepackage{cvpr}              

\definecolor{cvprblue}{rgb}{0.21,0.49,0.74}
\usepackage[pagebackref,breaklinks,colorlinks,allcolors=cvprblue]{hyperref}

\def\paperID{*****} 
\def\confName{CVPR}
\def\confYear{2025}

\title{Adaptive Contextual Embedding for Robust Far-View Borehole Detection}

\author{
Xuesong Liu$^{1}$ \quad Tianyu Hao$^{2}$ \quad Emmett J. Ientilucci$^{1}$\\[2mm]
$^{1}$Rochester Institute of Technology \quad $^{2}$Guangzhou University\\[1mm]
{\tt\small xl2088@rit.edu, howtyee@e.gzhu.edu.cn, ejipci@rit.edu}
}

\begin{document}
\maketitle
\begin{abstract}
In controlled blasting operations, accurately detecting densely distributed tiny boreholes from far-view imagery is critical for operational safety and efficiency. However, existing detection methods often struggle due to small object scales, highly dense arrangements, and limited distinctive visual features of boreholes. To address these challenges, we propose an adaptive detection approach that builds upon existing architectures (e.g., YOLO) by explicitly leveraging consistent embedding representations derived through exponential moving average (EMA)-based statistical updates.

Our method introduces three synergistic components: (1) adaptive augmentation utilizing dynamically updated image statistics to robustly handle illumination and texture variations; (2) embedding stabilization to ensure consistent and reliable feature extraction; and (3) contextual refinement leveraging spatial context for improved detection accuracy. The pervasive use of EMA in our method is particularly advantageous given the limited visual complexity and small scale of boreholes, allowing stable and robust representation learning even under challenging visual conditions. Experiments on a challenging proprietary quarry-site dataset demonstrate substantial improvements over baseline YOLO-based architectures, highlighting our method's effectiveness in realistic and complex industrial scenarios.
\end{abstract}    
\section{Introduction}
\label{sec:intro}

Rapid advancements in deep learning have significantly enhanced classical computer vision tasks such as image classification \cite{he2016deep, dosovitskiy2021image}, object detection \cite{ren2015faster, cai2018cascade, carion2020end}, and semantic segmentation \cite{liu2021swin, kirillov2023segment}. Despite these advances, specialized industrial settings—especially for detecting tiny, densely packed objects—remain challenging and relatively unexplored.

Industrial scenarios such as open-pit mining and quarry-site operations present unique visual complexities rarely seen in conventional detection datasets. Accurately identifying tiny boreholes used in controlled blasting requires robust handling of extremely small scales, dense object distributions, and substantial visual variability from textures, geological conditions, lighting, and environmental noise \cite{zhang2023mmpw, cui2024cpdd}. Standard detection methods frequently struggle under these conditions, causing elevated false alarms and missed targets, thus compromising operational efficiency and safety.

Existing techniques addressing these challenges typically include contextual integration \cite{chen2021contextnet}, adaptive data augmentation \cite{cubuk2020randaugment}, and embedding-based feature consistency methods \cite{wang2021dense}. However, these approaches have mostly been studied independently or in simpler scenarios, leaving their combined potential in realistic industrial contexts under-investigated.

To address these critical gaps, we propose an adaptive detection approach built upon established detection architectures (e.g., YOLO), specifically designed to detect densely distributed tiny boreholes reliably under complex quarry-site conditions. Our method leverages exponential moving averages (EMA) across multiple stages, ensuring robust learning given the small scale, dense distribution, and limited distinctive visual features of boreholes. Our key contributions include:

\begin{itemize}
\item \textbf{Adaptive Augmentation:} Dynamically updates augmentation parameters using EMA-based global image statistics, explicitly mitigating variability caused by diverse illumination conditions, soil textures, and geological backgrounds frequently encountered in borehole imagery.

\item \textbf{Embedding Stabilization:} Employs EMA-driven embedding updates to maintain stable and consistent feature extraction, explicitly addressing challenges posed by the small scale, limited distinctive visual features, and high-density distributions of boreholes, thus reducing false alarms and improving detection reliability.

\item \textbf{Contextual Refinement:} Integrates spatial context embeddings refined via EMA to effectively distinguish true borehole targets from visually ambiguous environmental noise such as dust, debris, and surface irregularities typical in quarry-site imagery.
\end{itemize}

Extensive experiments on a challenging proprietary quarry-site dataset demonstrate substantial improvements in detection accuracy and robustness when integrating our innovations into widely-adopted detection backbones such as YOLO-series \cite{bochkovskiy2020yolov4}, RCNN-series \cite{ren2015faster}, and NanoDet \cite{rangi2021nanodet}. Our approach thus advances both theoretical understanding and practical capabilities, significantly enhancing real-world operational effectiveness and safety.

\section{Method}
\label{sec:method}

We propose an adaptive detection method tailored for accurately detecting tiny, densely distributed boreholes in challenging industrial imagery. Given their small scale, dense arrangements, and limited distinctive features, we employ exponential moving average (EMA) techniques across our method to stabilize and enhance feature representations.

\begin{figure}[htbp]
    \centering
    \includegraphics[width=0.8\linewidth]{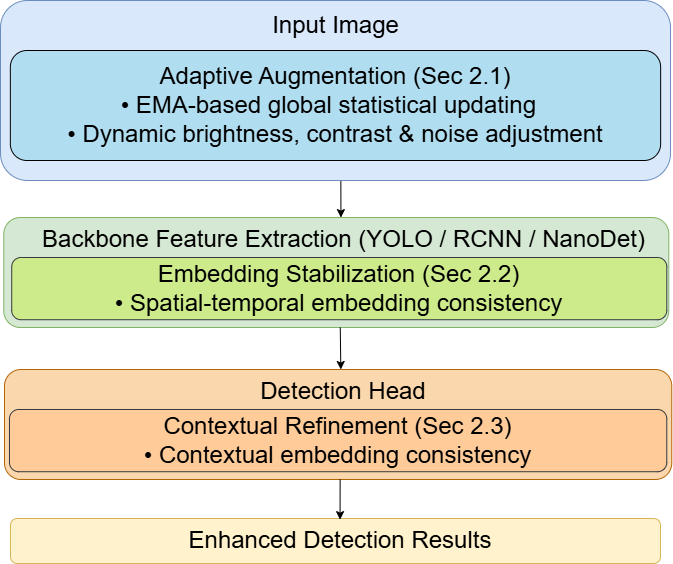}
    \caption{Overview of our adaptive detection method illustrating integration points of Adaptive Augmentation (Sec.~2.1), Embedding Stabilization (Sec.~2.2), and Contextual Refinement (Sec.~2.3).}
    \label{fig:flowchart}
\end{figure}

As illustrated in Fig.~\ref{fig:flowchart}, our approach integrates three key EMA-supported components: (1) \textbf{Adaptive Augmentation} to handle illumination and background variability; (2) \textbf{Embedding Stabilization} for consistent and robust feature extraction; and (3) \textbf{Contextual Refinement} to leverage spatial context, effectively distinguishing boreholes from visually similar background noise.

\subsection{Adaptive Augmentation}

Adaptive augmentation dynamically enhances borehole visibility and mitigates background variability (e.g., illumination, textures) using exponential moving averages (EMA). At iteration $t$, local ($\mu_{\Omega}^{(t)}, \sigma_{\Omega}^{(t)}$) and global ($\mu_{ref}^{(t)}, \sigma_{ref}^{(t)}$) image statistics are computed:
\begin{align}
\mu_{\Omega}^{(t)} &= \frac{1}{|\Omega|}\sum_{(u,v)\in \Omega}X^{(t)}(u,v), \\
\sigma_{\Omega}^{(t)} &= \sqrt{\frac{1}{|\Omega|}\sum_{(u,v)\in \Omega}(X^{(t)}(u,v)-\mu_{\Omega}^{(t)})^2},\\
\mu_{ref}^{(t)} &= (1-\rho)\mu_{ref}^{(t-1)} + \rho\,\mu_{batch}^{(t)}, \\
\sigma_{ref}^{(t)} &= (1-\rho)\sigma_{ref}^{(t-1)} + \rho\,\sigma_{batch}^{(t)},
\end{align}
where $\rho\in[0.01, 0.1]$ is the EMA smoothing factor, typically chosen to ensure gradual and stable statistical updates.

Augmentation parameters for brightness ($\alpha^{(t)}$), contrast ($\beta^{(t)}$), and additive noise ($\eta^{(t)}$) are adaptively derived as:
\begin{align}
\alpha^{(t)} &= 1 + k_1\frac{\mu_{ref}^{(t)} - \mu_{\Omega}^{(t)}}{\mu_{ref}^{(t)}}, \quad k_1\in[0.5, 1.5], \nonumber \\[4pt]
\beta^{(t)} &= 1 + k_2\frac{\sigma_{ref}^{(t)} - \sigma_{\Omega}^{(t)}}{\sigma_{ref}^{(t)}}, \quad k_2\in[0.5, 1.5], \nonumber \\[4pt]
\eta^{(t)} &= k_3\frac{|\sigma_{ref}^{(t)} - \sigma_{\Omega}^{(t)}|}{\sigma_{ref}^{(t)}}, \quad k_3\in[0.01, 0.1].
\end{align}

Here, the brightness ($\alpha^{(t)}$) and contrast ($\beta^{(t)}$) adjustments explicitly compensate for local deviations in illumination and texture compared to global references, enhancing feature consistency. The additive Gaussian noise ($\eta^{(t)}$) further enhances robustness against subtle background variations. The augmented image at iteration $t$ is computed as:
\small
\begin{equation}
\begin{aligned}
X'^{(t)}(u,v) &= \beta^{(t)}\alpha^{(t)}X^{(t)}(u,v)+\eta^{(t)}\epsilon(u,v),\\
\epsilon(u,v)&\sim\mathcal{N}(0,1).
\end{aligned}
\end{equation}

\subsection{Embedding Stabilization}

Embedding stabilization explicitly addresses embedding coherence challenges caused by small object scales, dense distributions, and limited visual distinctions among boreholes. At iteration $t$, embeddings $e_i^{(t)}$ are first spatially grouped into embedding sets $G_j^{(t)}$ based on spatial proximity:
\begin{equation}
G_j^{(t)}=\{e_i^{(t)}\mid \|c_i-c_j\|_2<\delta\},
\end{equation}
where $c_i, c_j$ denote embedding spatial centers, and proximity threshold $\delta \in [20,100]$ pixels determines local spatial coherence.

We progressively refine cluster-level ($\mu_j^{(t)}$) and global ($\mu_{global}^{(t)}$) embedding representations using exponential moving averages (EMA):
\begin{align}
\mu_j^{(t)} &= (1-\rho)\mu_j^{(t-1)} + \rho\frac{1}{|G_j^{(t)}|}\sum_{e\in G_j^{(t)}}e,\\[3pt]
\mu_{global}^{(t)} &= (1-\rho)\mu_{global}^{(t-1)}+\rho\frac{1}{N_t}\sum_j\sum_{e\in G_j^{(t)}}e,
\end{align}
where EMA smoothing factor $\rho\in[0.01,0.1]$ ensures stable updates.

Embedding consistency is enforced via two complementary losses: clustering consistency loss ($L_{cluster}^{(t)}$), ensuring spatial coherence of embeddings within each image, and stacking consistency loss ($L_{stack}^{(t)}$), enforcing temporal stability of embeddings across multiple images and training iterations:
\begin{equation}
\small
\resizebox{0.9\linewidth}{!}{$
L_{cluster}^{(t)} = \frac{1}{J_t}\sum_{j=1}^{J_t}\left(\frac{1}{|G_j^{(t)}|}\sum_{e\in G_j^{(t)}}\|e-\mu_j^{(t)}\|_2^2+\lambda\|\mu_j^{(t)}-\mu_{global}^{(t)}\|_2^2\right),
$}
\end{equation}
\begin{equation}
L_{stack}^{(t)}=\frac{1}{|G_j^{(t)}|}\sum_{e\in G_j^{(t)}}\|e-e_{stack}^{(t)}\|_2^2,
\end{equation}
where hyperparameter $\lambda\in[0.5,2.0]$ balances local cluster and global embedding coherence.

The stacked embedding representation ($e_{stack}^{(t)}$) aggregates embeddings temporally across training iterations:
\begin{equation}
e_{stack}^{(t)}=(1-\rho)e_{stack}^{(t-1)}+\rho\frac{1}{|G_j^{(t)}|}\sum_{e\in G_j^{(t)}}e.
\end{equation}

Thus, clustering consistency explicitly maintains spatial embedding coherence within each image, while stacking consistency explicitly ensures temporal embedding stability across image batches.

\subsection{Contextual Refinement}

Contextual refinement leverages spatial information around candidate detections, exploiting the observation that boreholes typically share embedding characteristics regardless of local density. We expand original bounding boxes $(x_i, y_i, w_i, h_i)$ by a factor $\gamma \in [0.1,0.5]$ to capture spatial context:
\begin{equation}
\tilde{b}_i^{(t)}=(x_i-\gamma w_i,\,y_i-\gamma h_i,\,w_i+2\gamma w_i,\,h_i+2\gamma h_i).
\end{equation}

Contextual embeddings $E_{context}^{(t)}(\tilde{b}_i)$ are extracted via ROI pooling from expanded bounding boxes and progressively refined using exponential moving averages (EMA):
\begin{equation}
E_{context}^{ref(t)}=(1-\rho)E_{context}^{ref(t-1)}+\rho\frac{1}{N_t}\sum_{i=1}^{N_t}E_{context}^{(t)}(\tilde{b}_i),
\end{equation}
where EMA smoothing $\rho\in[0.01,0.1]$ ensures stable updates.

We enforce consistency between the current contextual embeddings and their EMA-updated reference via a contextual consistency loss:
\begin{equation}
L_{context}^{(t)} = \frac{1}{N_t}\sum_{i=1}^{N_t}\|E_{context}^{(t)}(\tilde{b}_i)-E_{context}^{ref(t)}\|_2^2.
\end{equation}

These refined contextual embeddings are then concatenated with object embeddings $E_{object}^{(t)}(b_i)$, forming combined embeddings:
\begin{equation}
E_{merged}^{(t)}(b_i)=\text{Concat}(E_{object}^{(t)}(b_i), E_{context}^{ref(t)}).
\end{equation}

\subsection{Overall Objective}

Our adaptive detection approach integrates adaptive augmentation (Sec.~2.1), embedding stabilization (Sec.~2.2), and contextual refinement (Sec.~2.3) into standard detection architectures (e.g., YOLO). The total loss explicitly combines the standard detection losses with embedding and contextual consistency losses, each scaled by hyperparameters to control their influence:

\begin{equation}
L_{total}^{(t)} = L_{cls}^{(t)} + L_{bbox}^{(t)} + L_{obj}^{(t)} + \lambda_1 L_{cluster}^{(t)} + \lambda_2 L_{stack}^{(t)} + \lambda_3 L_{context}^{(t)},
\end{equation}
where:
\begin{itemize}
    \item $L_{cls}^{(t)}$: classification loss,
    \item $L_{bbox}^{(t)}$: bounding-box regression loss,
    \item $L_{obj}^{(t)}$: objectness confidence loss,
    \item $L_{cluster}^{(t)}$: spatial embedding clustering consistency loss,
    \item $L_{stack}^{(t)}$: temporal embedding stacking consistency loss,
    \item $L_{context}^{(t)}$: contextual embedding consistency loss,
    \item $\lambda_1, \lambda_2, \lambda_3$: weighting hyperparameters balancing detection accuracy and embedding/contextual coherence, typically chosen from $[0.1,1.0]$.
\end{itemize}

These auxiliary embedding and contextual consistency losses explicitly regularize the embedding space and contextual representations, directly influencing the learned features and predictions, while the original detection loss structure ensures compatibility with established detection architectures.

\section{Experiments}
\label{sec:4_experiments}

We evaluate our adaptive detection approach on a challenging proprietary dataset collected from multiple quarry sites. We quantitatively and qualitatively assess the effectiveness of each component: Adaptive Augmentation (AA), Embedding Stabilization (ES), and Contextual Refinement (CR).

\subsection{Dataset and Experimental Setup}

Our proprietary dataset consists of 250 annotated images, each containing an average of 100 densely distributed boreholes. Images feature substantial variability in illumination, textures, resolution, and noise levels, realistically representing typical quarry-site conditions.

The detection model is implemented in PyTorch, using a YOLO-based architecture, and trained for 100 epochs on an NVIDIA GeForce RTX 4090 GPU. We optimize the model using AdamW with a learning rate of $1\times10^{-3}$ and a batch size of 8. Performance is assessed using standard detection metrics: mean Average Precision (mAP), Precision, Recall, and F1-score.

\subsection{Effectiveness of Proposed Components}
\begin{figure*}[htbp]
    \centering
    \includegraphics[width=2\columnwidth]{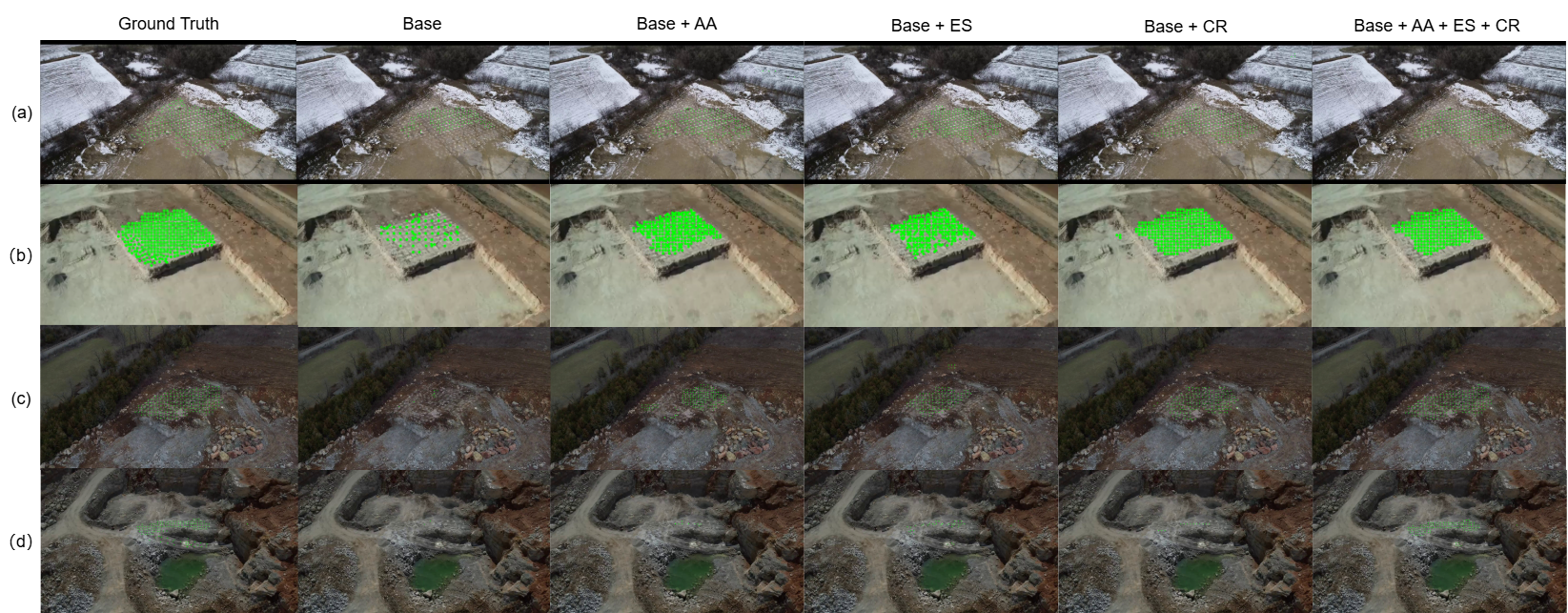}
    \caption{Qualitative comparison of detection results: Ground Truth, Baseline YOLO (Base), and incremental addition of proposed components. Our complete model (AA + ES + CR) shows robust and accurate detections, especially in challenging conditions.}
    
    \label{fig:qualitative_results}
\end{figure*}
Due to the limited size of our dataset, we employed k-fold cross-validation to robustly evaluate the effectiveness of each proposed component. Table~\ref{table:ablation_study} presents the mean and standard deviation of the evaluation metrics across the folds.

To systematically analyze the contributions of our components, we first compared results from a baseline YOLO model (YOLOv11) without our components. Subsequently, we added each proposed component individually, then combined them in pairs, and finally incorporated all three components together, as shown in Table~\ref{table:ablation_study}. From the table, we observe clear incremental improvements: adding a single component noticeably enhances performance, while adding multiple components leads to progressively better results. Integrating all three components (AA+ES+CR) achieves the highest overall performance across all metrics.

\begin{table}[htbp]
\centering
\caption{Component-wise evaluation (k-fold cross-validation).}
\label{table:ablation_study}
\small
\setlength{\tabcolsep}{2pt}
\begin{tabular}{lcccc}
\toprule
\textbf{Model} & \textbf{mAP (\%)} & \textbf{Precision (\%)} & \textbf{Recall (\%)} & \textbf{F1 (\%)} \\
\midrule
Baseline & 61.5±0.8 & 59.3±0.7 & 62.1±0.9 & 60.7±0.8 \\
+ AA\textsuperscript{*} & 64.5±0.6 & 61.8±0.5 & 65.2±0.7 & 63.5±0.6 \\
+ ES\textsuperscript{†} & 66.8±0.7 & 64.0±0.5 & 67.8±0.6 & 65.8±0.6 \\
+ CR\textsuperscript{‡} & 67.5±0.7 & 64.7±0.6 & 68.3±0.8 & 66.5±0.7 \\
+ AA + ES & 70.3±0.6 & 67.3±0.4 & 71.5±0.5 & 69.3±0.5 \\
+ AA + CR & 70.1±0.6 & 67.0±0.4 & 71.0±0.5 & 68.9±0.4 \\
+ ES + CR & 72.5±0.5 & 69.8±0.4 & 73.2±0.5 & 71.5±0.4 \\
+ All & \textbf{74.9±0.4} & \textbf{72.8±0.3} & \textbf{75.6±0.4} & \textbf{74.2±0.3} \\
\bottomrule
\end{tabular}
\vspace{1mm}
\footnotesize{
\textsuperscript{*}AA: Adaptive Augmentation, 
\textsuperscript{†}ES: Embedding Stabilization, 
\textsuperscript{‡}CR: Contextual Refinement.
}
\vspace{2mm}
\end{table}

We further illustrate qualitative improvements using selected visual results, shown in Fig.~\ref{fig:qualitative_results}. Due to the tiny size and high density of boreholes, we denote detections with green dots instead of bounding boxes for clearer visualization. Four representative images with diverse conditions are presented: Fig.~\ref{fig:qualitative_results}(a) shows a snowy winter blasting scenario, challenging due to the color similarity between snow and crushed rock particles around boreholes. Fig.~\ref{fig:qualitative_results}(b) demonstrates a lower-resolution image scenario, which often occurs due to equipment limitations at quarry sites. Fig.~\ref{fig:qualitative_results}(c) represents relatively dark illumination and an uneven bench surface with structural and textural complexity. Finally, Fig.~\ref{fig:qualitative_results}(d) presents the most challenging scenario, featuring dark lighting conditions, uneven surfaces, water pools, and boreholes positioned across different quarry levels.

These visual examples highlight specific contributions of each component. The baseline method struggles significantly, exhibiting numerous missed detections (false negatives), particularly evident in Fig.~\ref{fig:qualitative_results}(c) and (d). Incorporating adaptive augmentation (AA) reduces missed detections by stabilizing the model's response to varying environmental conditions, clearly visible in Fig.~\ref{fig:qualitative_results}(a) and (b). Embedding stabilization (ES) further improves completeness, increasing correctly identified boreholes across scenarios. Contextual refinement (CR) notably enhances spatial accuracy and reduces noise-induced false positives. Ultimately, the fully integrated model (AA+ES+CR) achieves robust and accurate detections, even under the most challenging conditions.

\section{Conclusion}
\label{sec:5_conclusion}
In this paper, we introduced a robust detection approach leveraging adaptive augmentation, embedding stabilization, and contextual refinement, significantly enhancing borehole detection accuracy and robustness under complex quarry-site conditions. Our method effectively reduces false detections and ensures reliable identification, even amid challenging environmental variations and ambiguous visual cues.

Despite these advancements, our approach currently relies on existing detection frameworks such as YOLOv11, involving substantial computational resources due to their large parameter counts. Designing a dedicated, efficient detection framework specifically tailored for dense, small-object detection scenarios represents a promising direction for future research, potentially achieving comparable performance with significantly reduced model complexity. Additionally, further investigations could explore applying our context-aware embedding strategies within advanced, transformer-based detection architectures, and validating the generalization of our method across broader industrial contexts, including construction sites and diverse geological settings.

\clearpage
{
    \small
    \bibliographystyle{ieeenat_fullname}
    \bibliography{main}

\begin{thebibliography}{28}
\providecommand{\natexlab}[1]{#1}
\providecommand{\url}[1]{\texttt{#1}}
\expandafter\ifx\csname urlstyle\endcsname\relax
  \providecommand{\doi}[1]{doi: #1}\else
  \providecommand{\doi}{doi: \begingroup \urlstyle{rm}\Url}\fi

\bibitem[Bochkovskiy et~al.(2020)Bochkovskiy, Wang, and Liao]{bochkovskiy2020yolov4}
Alexey Bochkovskiy, Chien-Yao Wang, and Hong-Yuan~Mark Liao.
\newblock {YOLOv4}: Optimal speed and accuracy of object detection.
\newblock In \emph{Proceedings of the IEEE/CVF Conference on Computer Vision and Pattern Recognition (CVPR)}, pages 10934--10943, 2020.

\bibitem[Cai and Vasconcelos(2018)]{cai2018cascade}
Zhaowei Cai and Nuno Vasconcelos.
\newblock Cascade r-cnn: Delving into high quality object detection.
\newblock In \emph{Proceedings of the IEEE Conference on Computer Vision and Pattern Recognition}, pages 6154--6162, 2018.

\bibitem[Cao et~al.(2019)Cao, Xu, Lin, Wei, and Hu]{cao2019gcnet}
Yue Cao, Jiarui Xu, Stephen Lin, Fangyun Wei, and Han Hu.
\newblock Gcnet: Non-local networks meet squeeze-excitation networks and beyond.
\newblock In \emph{Proceedings of the IEEE/CVF International Conference on Computer Vision Workshops (ICCVW)}, pages 1971--1980, 2019.

\bibitem[Carion et~al.(2020)Carion, Massa, Synnaeve, Usunier, Kirillov, and Zagoruyko]{carion2020end}
Nicolas Carion, Francisco Massa, Gabriel Synnaeve, Nicolas Usunier, Alexander Kirillov, and Sergey Zagoruyko.
\newblock End-to-end object detection with transformers.
\newblock In \emph{European Conference on Computer Vision}, pages 213--229. Springer, 2020.

\bibitem[Chen et~al.(2020)Chen, Liu, Zhao, and Jia]{chen2020gridmask}
Pengguang Chen, Shu Liu, Hengshuang Zhao, and Jiaya Jia.
\newblock Gridmask data augmentation.
\newblock In \emph{Proceedings of the IEEE/CVF Conference on Computer Vision and Pattern Recognition (CVPR)}, pages 117--126, 2020.

\bibitem[Chen et~al.(2021)Chen, Chen, Lin, See, Yu, and Ke]{chen2021contextnet}
Zhiming Chen, Kean Chen, Weiyao Lin, John See, Hui Yu, and Yan Ke.
\newblock Contextnet: A general framework for object detection with context information.
\newblock \emph{Pattern Recognition}, 115:\penalty0 107883, 2021.

\bibitem[Chen et~al.(2023)Chen, Lu, Wang, Zhang, and Zhou]{chen2023cross}
Zhe Chen, Xunyu Lu, Yiran Wang, Hao Zhang, and Wei Zhou.
\newblock Cross-scale attention for small object detection.
\newblock In \emph{Proceedings of the IEEE/CVF Winter Conference on Applications of Computer Vision (WACV)}, pages 3627--3636, 2023.

\bibitem[Cubuk et~al.(2019)Cubuk, Zoph, Mane, Vasudevan, and Le]{cubuk2019autoaugment}
Ekin~D. Cubuk, Barret Zoph, Dandelion Mane, Vijay Vasudevan, and Quoc~V. Le.
\newblock Autoaugment: Learning augmentation policies from data.
\newblock \emph{Proceedings of the IEEE/CVF Conference on Computer Vision and Pattern Recognition}, pages 113--123, 2019.

\bibitem[Cubuk et~al.(2020)Cubuk, Zoph, Shlens, and Le]{cubuk2020randaugment}
Ekin~D. Cubuk, Barret Zoph, Jonathon Shlens, and Quoc~V. Le.
\newblock Randaugment: Practical automated data augmentation with a reduced search space.
\newblock In \emph{Proceedings of the IEEE/CVF Conference on Computer Vision and Pattern Recognition Workshops}, pages 702--703, 2020.

\bibitem[Cui et~al.(2024)Cui, Yan, Yu, Wang, and Liu]{cui2024cpdd}
Xiangjie Cui, Shitong Yan, Hongkai Yu, Yang Wang, and Jiangbo Liu.
\newblock {CPDD-YOLOv8: A Robust Detector for Tiny Objects in Aerial Images}.
\newblock \emph{IEEE Transactions on Geoscience and Remote Sensing}, 62:\penalty0 1--15, 2024.

\bibitem[Dosovitskiy et~al.(2021)Dosovitskiy, Beyer, Kolesnikov, Weissenborn, Zhai, Unterthiner, Dehghani, Minderer, Heigold, Gelly, Uszkoreit, and Houlsby]{dosovitskiy2021image}
Alexey Dosovitskiy, Lucas Beyer, Alexander Kolesnikov, Dirk Weissenborn, Xiaohua Zhai, Thomas Unterthiner, Mostafa Dehghani, Matthias Minderer, Georg Heigold, Sylvain Gelly, Jakob Uszkoreit, and Neil Houlsby.
\newblock An image is worth 16x16 words: Transformers for image recognition at scale.
\newblock In \emph{International Conference on Learning Representations}, 2021.

\bibitem[Ge et~al.(2021)Ge, Liu, Wang, Li, and Sun]{ge2021yolox}
Zheng Ge, Songtao Liu, Feng Wang, Zeming Li, and Jian Sun.
\newblock Yolox: Exceeding yolo series in 2021.
\newblock \emph{arXiv preprint arXiv:2107.08430}, 2021.

\bibitem[Ghiasi et~al.(2021)Ghiasi, Cui, Srinivas, Qian, Lin, Cubuk, Le, and Zoph]{ghiasi2021simple}
Golnaz Ghiasi, Yin Cui, Aravind Srinivas, Rui Qian, Tsung-Yi Lin, Ekin~D. Cubuk, Quoc~V. Le, and Barret Zoph.
\newblock Simple copy-paste is a strong data augmentation method for instance segmentation.
\newblock In \emph{Proceedings of the IEEE/CVF Conference on Computer Vision and Pattern Recognition (CVPR)}, pages 2918--2928, 2021.

\bibitem[He et~al.(2016)He, Zhang, Ren, and Sun]{he2016deep}
Kaiming He, Xiangyu Zhang, Shaoqing Ren, and Jian Sun.
\newblock Deep residual learning for image recognition.
\newblock In \emph{Proceedings of the IEEE Conference on Computer Vision and Pattern Recognition}, pages 770--778, 2016.

\bibitem[Kirillov et~al.(2023)Kirillov, Mintun, Ravi, Mao, Rolland, Gustafson, Xiao, Whitehead, Berg, Lo, Doll{\'a}r, and Girshick]{kirillov2023segment}
Alexander Kirillov, Eric Mintun, Nikhila Ravi, Hanzi Mao, Chloe Rolland, Laura Gustafson, Tete Xiao, Spencer Whitehead, Alexander~C Berg, Wan-Yen Lo, Piotr Doll{\'a}r, and Ross Girshick.
\newblock Segment anything.
\newblock In \emph{Proceedings of the IEEE/CVF International Conference on Computer Vision}, pages 20942--20951, 2023.

\bibitem[Liu et~al.(2018)Liu, Qi, Qin, Shi, and Jia]{liu2018path}
Shu Liu, Lu Qi, Haifang Qin, Jianping Shi, and Jiaya Jia.
\newblock Path aggregation network for instance segmentation.
\newblock In \emph{Proceedings of the IEEE Conference on Computer Vision and Pattern Recognition (CVPR)}, pages 8759--8768, 2018.

\bibitem[Liu et~al.(2019)Liu, Huang, and Wang]{liu2019learning}
Songtao Liu, Di Huang, and Yunhong Wang.
\newblock Learning adaptive spatial fusion for object detection.
\newblock In \emph{Proceedings of the IEEE/CVF Conference on Computer Vision and Pattern Recognition (CVPR)}, pages 11462--11471, 2019.

\bibitem[Liu et~al.(2021)Liu, Lin, Cao, Hu, Wei, Zhang, Lin, and Guo]{liu2021swin}
Ze Liu, Yutong Lin, Yue Cao, Han Hu, Yixuan Wei, Zheng Zhang, Stephen Lin, and Baining Guo.
\newblock Swin transformer: Hierarchical vision transformer using shifted windows.
\newblock In \emph{Proceedings of the IEEE/CVF International Conference on Computer Vision}, pages 10012--10022, 2021.

\bibitem[RangiLyu(2021)]{rangi2021nanodet}
RangiLyu.
\newblock Nanodet: Tiny object detection for resource-constrained devices.
\newblock \emph{arXiv preprint arXiv:2103.13744}, 2021.

\bibitem[Ren et~al.(2015)Ren, He, Girshick, and Sun]{ren2015faster}
Shaoqing Ren, Kaiming He, Ross Girshick, and Jian Sun.
\newblock Faster r-cnn: Towards real-time object detection with region proposal networks.
\newblock In \emph{Advances in Neural Information Processing Systems}, pages 91--99, 2015.

\bibitem[Roh et~al.(2022)Roh, Shin, Kim, and Kim]{roh2022sparse}
Byungseok Roh, Junyong Shin, Wuhyun Kim, and Saehoon Kim.
\newblock Sparse detr: Efficient end-to-end object detection with learnable sparsity.
\newblock In \emph{International Conference on Learning Representations (ICLR)}, 2022.

\bibitem[Tan et~al.(2020)Tan, Pang, and Le]{tan2020efficientdet}
Mingxing Tan, Ruoming Pang, and Quoc~V. Le.
\newblock Efficientdet: Scalable and efficient object detection.
\newblock In \emph{Proceedings of the IEEE/CVF Conference on Computer Vision and Pattern Recognition (CVPR)}, pages 10781--10790, 2020.

\bibitem[Wang et~al.(2021)Wang, Gao, Li, Yu, and Xiang]{wang2021dense}
Jinpeng Wang, Yuting Gao, Ke Li, Yifan Yu, and Tao Xiang.
\newblock Dense embedding contrast for unsupervised semantic segmentation.
\newblock In \emph{Proceedings of the IEEE/CVF Conference on Computer Vision and Pattern Recognition}, pages 8358--8367, 2021.

\bibitem[Wang et~al.(2018)Wang, Girshick, Gupta, and He]{wang2018non}
Xiaolong Wang, Ross Girshick, Abhinav Gupta, and Kaiming He.
\newblock Non-local neural networks.
\newblock In \emph{Proceedings of the IEEE Conference on Computer Vision and Pattern Recognition (CVPR)}, pages 7794--7803, 2018.

\bibitem[Yun et~al.(2019)Yun, Han, Oh, Chun, Choe, and Yoo]{yun2019cutmix}
Sangdoo Yun, Dongyoon Han, Seong~Joon Oh, Sanghyuk Chun, Junsuk Choe, and Youngjoon Yoo.
\newblock Cutmix: Regularization strategy to train strong classifiers with localizable features.
\newblock In \emph{Proceedings of the IEEE/CVF International Conference on Computer Vision (ICCV)}, pages 6023--6032, 2019.

\bibitem[Zhang et~al.(2018)Zhang, Cisse, Dauphin, and Lopez-Paz]{zhang2017mixup}
Hongyi Zhang, Moustapha Cisse, Yann~N. Dauphin, and David Lopez-Paz.
\newblock Mixup: Beyond empirical risk minimization.
\newblock In \emph{International Conference on Learning Representations (ICLR)}, 2018.

\bibitem[Zhang et~al.(2023)Zhang, Hou, Cui, and Zhang]{zhang2023mmpw}
Yuxin Zhang, Jie Hou, Haochen Cui, and Xiangrong Zhang.
\newblock {MMPW-Net: A Novel Detection Network for Tiny Objects in Remote Sensing Images Based on Mixed Minimum Point-Wasserstein Distance}.
\newblock \emph{Remote Sensing}, 15\penalty0 (19):\penalty0 4728, 2023.

\bibitem[Zhu et~al.(2021)Zhu, Su, Lu, Li, Wang, and Dai]{zhu2020deformable}
Xizhou Zhu, Weijie Su, Lewei Lu, Bin Li, Xiaogang Wang, and Jifeng Dai.
\newblock Deformable detr: Deformable transformers for end-to-end object detection.
\newblock In \emph{International Conference on Learning Representations (ICLR)}, 2021.

\end{thebibliography}
}

\clearpage
\setcounter{page}{1}
\maketitlesupplementary

\section{Component-wise Comparisons}
\label{sec:supp_components}

\subsection{Adaptive Augmentation (AA)}
\label{sec:supp_aa}

Quantitative comparisons provided in Table~\ref{tab:aa_comparison} demonstrate that our AA method alone does not significantly outperform other augmentation techniques, suggesting that adaptive augmentation contributes only modestly when applied independently. However, when combined with all three proposed components, our method achieves substantial overall performance improvements.

\begin{table}[h]
\centering
\caption{Comparison of Adaptive Augmentation (AA) with other augmentation methods on YOLOv11 backbone.}
\label{tab:aa_comparison}
\scriptsize
\setlength{\tabcolsep}{1.5pt}
\begin{tabular}{lccccc}
\toprule
\textbf{Method} & \textbf{Year} & \textbf{mAP (\%)} & \textbf{Precision (\%)} & \textbf{Recall (\%)} & \textbf{F1 (\%)} \\
\midrule
MixUp~\cite{zhang2017mixup} & 2017 & 62.8±0.8 & 60.6±0.7 & 63.5±0.8 & 62.0±0.7 \\
CutMix~\cite{yun2019cutmix} & 2019 & 63.0±0.7 & 60.9±0.6 & 63.7±0.7 & 62.3±0.6 \\
AutoAugment~\cite{cubuk2019autoaugment} & 2019 & 63.4±0.7 & 61.2±0.6 & 64.0±0.7 & 62.6±0.6 \\
Mosaic~\cite{bochkovskiy2020yolov4} & 2020 & 62.7±0.7 & 60.5±0.6 & 63.3±0.7 & 61.9±0.7 \\
RandAugment~\cite{cubuk2020randaugment} & 2020 & 63.6±0.7 & 61.4±0.6 & 64.2±0.7 & 62.8±0.6 \\
GridMask~\cite{chen2020gridmask} & 2020 & 63.8±0.7 & 61.6±0.6 & 64.3±0.7 & 62.9±0.6 \\
Copy-Paste~\cite{ghiasi2021simple} & 2021 & 64.1±0.7 & 61.8±0.6 & 64.6±0.7 & 63.2±0.6 \\
\midrule
Ours (AA) & 2025 & \textbf{64.5±0.6} & \textbf{61.8±0.5} & \textbf{65.2±0.7} & \textbf{63.5±0.6} \\
Ours (Full Method) & 2025 & \textbf{74.9±0.4} & \textbf{72.8±0.3} & \textbf{75.6±0.4} & \textbf{74.2±0.3} \\
\bottomrule
\end{tabular}
\vspace{2mm}
\end{table}

\subsection{Embedding Stabilization (ES)}
\label{sec:supp_sc}

Our Embedding Stabilization (ES) technique leverages exponential moving averages to maintain consistent and robust embedding representations, effectively mitigating feature variations caused by challenging conditions such as dense object distributions and limited visual distinctions in quarry-site imagery.

Table~\ref{tab:sc_comparison} provides a quantitative comparison of ES against popular multi-scale feature fusion methods, including PANet, BiFPN, ASFF, and PAFPN. The results indicate that our ES approach achieves moderate but consistent performance improvements in mAP, Precision, Recall, and F1-score compared to these established methods. While these improvements demonstrate that embedding stabilization alone contributes modestly, it significantly enhances performance when combined with our complete set of proposed components. This clearly highlights the effectiveness of ES in providing complementary embedding consistency within our comprehensive detection approach.

\begin{table}[h]
\centering
\caption{Comparison of Embedding Stabilization (ES) with other feature fusion methods (YOLOv11 backbone).}
\label{tab:sc_comparison}
\scriptsize
\setlength{\tabcolsep}{2pt}
\begin{tabular}{lccccc}
\toprule
\textbf{Method} & \textbf{Year} & \textbf{mAP (\%)} & \textbf{Precision (\%)} & \textbf{Recall (\%)} & \textbf{F1 (\%)} \\
\midrule
PANet~\cite{liu2018path} & 2018 & 63.2±0.7 & 60.8±0.6 & 63.7±0.8 & 62.2±0.7 \\
BiFPN~\cite{tan2020efficientdet} & 2020 & 64.5±0.6 & 62.0±0.5 & 65.0±0.7 & 63.5±0.6 \\
ASFF~\cite{liu2019learning} & 2019 & 64.0±0.7 & 61.7±0.6 & 64.6±0.7 & 63.1±0.7 \\
PAFPN~\cite{ge2021yolox} & 2021 & 64.2±0.6 & 61.8±0.5 & 64.8±0.7 & 63.2±0.6 \\
\midrule
Ours (ES) & 2025 & \textbf{66.8±0.7} & \textbf{64.0±0.5} & \textbf{67.8±0.6} & \textbf{65.8±0.6} \\
Ours (Full Method) & 2025 & \textbf{74.9±0.4} & \textbf{72.8±0.3} & \textbf{75.6±0.4} & \textbf{74.2±0.3} \\
\bottomrule
\end{tabular}
\vspace{2mm}
\end{table}

\subsection{Contextual Refinement (CR)}
\label{sec:supp_ci}

Our Contextual Refinement (CR) component explicitly integrates spatial context into detection predictions through refined embedding representations. By leveraging exponential moving averages (EMA) of contextual embeddings, CR effectively enhances the discrimination between actual borehole targets and visually similar background noise, which is critical in dense and visually challenging quarry-site scenarios.

Table~\ref{tab:ci_comparison} quantitatively compares our CR method with existing state-of-the-art context-aware techniques, such as Non-local, GCNet, Transformer, and Deformable Attention. The results demonstrate that CR consistently delivers moderate but notable improvements across all key detection metrics (mAP, Precision, Recall, and F1-score). While the individual performance gains of CR are modest, it significantly contributes to the overall performance when combined with Adaptive Augmentation and Embedding Stabilization, highlighting its valuable role in our complete detection approach.

\begin{table}[h]
\centering
\caption{Comparison of Contextual Refinement (CR) with other contextual methods on YOLOv11 backbone.}
\label{tab:ci_comparison}
\scriptsize
\setlength{\tabcolsep}{1.5pt}
\begin{tabular}{lccccc}
\toprule
\textbf{Method} & \textbf{Year} & \textbf{mAP (\%)} & \textbf{Precision (\%)} & \textbf{Recall (\%)} & \textbf{F1 (\%)} \\
\midrule
Non-local~\cite{wang2018non} & 2018 & 63.8±0.7 & 61.2±0.6 & 64.5±0.7 & 62.8±0.6 \\
GCNet~\cite{cao2019gcnet} & 2019 & 64.3±0.6 & 61.7±0.6 & 65.0±0.6 & 63.3±0.6 \\
Transformer~\cite{carion2020end} & 2020 & 65.2±0.6 & 62.5±0.5 & 65.9±0.6 & 64.1±0.5 \\
Deformable Attention~\cite{zhu2020deformable} & 2020 & 65.7±0.5 & 63.0±0.5 & 66.4±0.5 & 64.6±0.5 \\
Sparse DETR~\cite{roh2022sparse} & 2022 & 65.9±0.5 & 63.3±0.4 & 66.7±0.5 & 64.9±0.5 \\
Cross-scale Attention~\cite{chen2023cross} & 2023 & 66.3±0.5 & 63.7±0.4 & 67.1±0.5 & 65.3±0.4 \\
\midrule
Ours (CR) & 2025 & \textbf{67.5±0.7} & \textbf{64.7±0.6} & \textbf{68.3±0.8} & \textbf{66.5±0.7} \\
Ours (Full Method) & 2025 & \textbf{74.9±0.4} & \textbf{72.8±0.3} & \textbf{75.6±0.4} & \textbf{74.2±0.3} \\
\bottomrule
\end{tabular}
\vspace{2mm}
\end{table}
\section{Model Size Sensitivity Analysis}
\label{sec:model_size}

\begin{table}[htbp]
\centering
\caption{Model size sensitivity analysis of YOLOv11 backbone variants (our method).}
\label{tab:model_size_sensitivity}
\scriptsize
\setlength{\tabcolsep}{1pt}
\begin{tabular}{lccccc}
\toprule
\textbf{Backbone} & \textbf{Params (M)} & \textbf{mAP (\%)} & \textbf{Precision (\%)} & \textbf{Recall (\%)} & \textbf{F1-score (\%)} \\
\midrule
Large & 72.5 & \textbf{74.9±0.4} & \textbf{72.8±0.3} & \textbf{75.6±0.4} & \textbf{74.2±0.3} \\
Medium & 36.7 & 71.3±0.5 & 69.2±0.4 & 72.0±0.5 & 70.6±0.4 \\
Small & 18.2 & 67.8±0.5 & 65.4±0.5 & 68.5±0.6 & 66.9±0.5 \\
Nano & 9.1 & 63.2±0.6 & 60.9±0.5 & 64.0±0.6 & 62.4±0.6 \\
\bottomrule
\end{tabular}
\vspace{2mm}
\end{table}

To assess how model size affects detection performance, we evaluated several YOLOv11 backbone variants (Large, Medium, Small, and Nano). Table~\ref{tab:model_size_sensitivity} clearly shows that detection accuracy consistently decreases as model complexity (number of parameters) is reduced. Specifically, shrinking the backbone from Large (72.5M parameters) to Nano (9.1M parameters) leads to an 11.7\% drop in mAP, emphasizing the importance of sufficient model capacity for accurate detection in challenging quarry-site conditions.

%
\begin{table*}[t!]
\centering
\caption{Performance comparison between original YOLO-series models and our enhanced versions.}
\label{table:yolo_comparison}
\scriptsize
\setlength{\tabcolsep}{5pt}
\begin{tabular}{lcccccccc}
\toprule
{\textbf{Model}} & \multicolumn{2}{c}{\textbf{mAP (\%)}} & \multicolumn{2}{c}{\textbf{Precision (\%)}} & \multicolumn{2}{c}{\textbf{Recall (\%)}} & \multicolumn{2}{c}{\textbf{F1-score (\%)}} \\
\cmidrule(lr){2-3} \cmidrule(lr){4-5} \cmidrule(lr){6-7} \cmidrule(lr){8-9}
 & Original & Enhanced & Original & Enhanced & Original & Enhanced & Original & Enhanced \\
\midrule
YOLOv3 & 55.7±0.8 & \textbf{65.3±0.7} & 53.8±0.7 & \textbf{63.2±0.6} & 56.1±0.8 & \textbf{66.1±0.7} & 54.9±0.7 & \textbf{64.6±0.6} \\
YOLOv6 & 60.2±0.7 & \textbf{70.9±0.6} & 58.0±0.6 & \textbf{68.4±0.5} & 60.9±0.7 & \textbf{71.6±0.5} & 59.4±0.6 & \textbf{69.9±0.5} \\
YOLOv7 & 62.1±0.6 & \textbf{73.8±0.5} & 60.1±0.5 & \textbf{71.3±0.4} & 62.7±0.6 & \textbf{74.5±0.4} & 61.4±0.5 & \textbf{72.9±0.4} \\
YOLOv8 & 63.7±0.5 & \textbf{75.9±0.4} & 61.8±0.5 & \textbf{73.7±0.3} & 64.2±0.5 & \textbf{76.5±0.4} & 63.0±0.5 & \textbf{75.1±0.3} \\
YOLOv9 & 65.4±0.5 & \textbf{78.2±0.4} & 63.6±0.4 & \textbf{75.9±0.3} & 65.9±0.5 & \textbf{78.8±0.3} & 64.7±0.4 & \textbf{77.3±0.3} \\
YOLOv11 & 61.5±0.8 & \textbf{74.9±0.4} & 59.3±0.7 & \textbf{72.8±0.3} & 62.1±0.9 & \textbf{75.6±0.4} & 60.7±0.8 & \textbf{74.2±0.3} \\
\bottomrule
\end{tabular}
\vspace{2mm}
\end{table*}

\section{Generalization Across Different YOLO Backbones}

To assess the generalizability and robustness of our proposed approach, we evaluated its performance across multiple YOLO-series backbones, ranging from earlier (YOLOv3) to recent architectures (YOLOv11). Table~\ref{table:yolo_comparison} clearly illustrates that our enhanced model consistently achieves substantial improvements across all key detection metrics (mAP, Precision, Recall, and F1-score) compared to their original implementations. 

Particularly, when applied to the most recent YOLOv11 backbone, our approach significantly increases the mean Average Precision (mAP) from 61.5\% to 74.9\%, demonstrating the effectiveness of our adaptive augmentation, embedding stabilization, and contextual refinement components in complex quarry-site scenarios. Similar performance gains are observed consistently across earlier YOLO versions, highlighting the general applicability and versatility of our proposed techniques across different YOLO-based detection frameworks.

\end{document}